# Physics-Informed Extreme Learning Machine (PIELM) for Tunnelling-Induced Soil-Pile Interactions

Fu-Chen **Guo**, PhD student
Email: guofuchen@mail.sdu.edu.cn
School of Qilu Transportation, Shandong University, Jinan, 250002, China

Professor Pei-Zhi **Zhuang**
E-mail: zhuangpeizhi@sdu.edu.cn
School of Qilu Transportation, Shandong University, Jinan, 250002, China

Dr Fei **Ren**
Email: ren87@outlook.com
School of Mechanical Engineering, Shandong Key Laboratory of CNC Machine Tool Functional Components, Qilu University of Technology, Jinan, 250353, China

Dr Hong-Ya **Yue**
Email: yuehongya@mail.sdu.edu.cn
School of Qilu Transportation, Shandong University, Jinan, 250002, China

Dr He **Yang**
Email: yanghesdu@mail.sdu.edu.cn
Corresponding author
School of Qilu Transportation, Shandong University, Jinan, 250002, China

**Physics-Informed Extreme Learning Machine (PIELM) for Tunnelling-Induced Soil-Pile Interactions**

Authors: Fu-Chen **Guo**, Pei-Zhi **Zhuang**, Fei **Ren**, Hong-Ya **Yue**, He **Yang**

## Abstract

Physics-informed machine learning has been a promising data-driven and physics-informed approach in geotechnical engineering. This study proposes a physics-informed extreme learning machine (PIELM) framework for analyzing tunneling-induced soil-pile interactions. The pile foundation is modeled as an Euler-Bernoulli beam, and the surrounding soil is modeled as a Pasternak foundation. The soil-pile interaction is formulated into a fourth-order ordinary differential equation (ODE) that constitutes the physics-informed component, while measured data are incorporated into PIELM as the data-driven component. Combining physics and data yields a loss vector of the extreme learning machine (ELM) network, which is trained within 1 second by the least squares method. After validating the PIELM approach by the boundary element method (BEM) and finite difference method (FDM), parametric studies are carried out to examine the effects of ELM network architecture, data monitoring locations and numbers on the performance of PIELM. The results indicate that monitored data should be placed at positions where the gradients of pile deflections are significant, such as at the pile tip/top and near tunneling zones. Two application examples highlight the critical role of physics-informed and data-driven approach for tunnelling-induced soil-pile interactions. The proposed approach shows great potential for real-time monitoring and safety assessment of pile foundations, and benefits for intelligent early-warning systems in geotechnical engineering.

**Keywords:** physics-informed machine learning, physics-informed neural networks, physics-informed extreme learning machine, soil-pile interactions, tunnelling, pile foundation

# 1. Introduction

Piles are widely used as foundations of buildings and infrastructures. With the acceleration of global urbanization, tunnelling for the construction of subways and utility tunnels inevitably and frequently encounters scenarios where tunnels pass beneath piles (Son and Moorak 2015; Bilotta et al. 2017; Yang et al. 2025d). The soil disturbances caused by tunneling alter the original stress equilibrium and may exert significant effects on nearby pile foundations. Since minor deformation of piles may threaten the stability and serviceability of superstructures, understanding and monitoring soil-pile interactions induced by tunneling are crucial for risk management in urban underground constructions.

Various approaches have been reported to analyze tunnelling-induced soil-pile interactions, including field and centrifuge model tests (Lee and Bassett 2007; Ng and Lu 2014; Soomro et al. 2020), theoretical solutions (Liu et al. 2018), and numerical simulations (Xu and Poulos 2001; Basile 2014). Experimental tests provide valuable data that capture the actual mechanical responses of piles during tunnel excavation and serve as an important basis for discovering and validating physical laws. The primary limitation is that the amount of such data is often constrained by the high costs associated with testing time, space, and budget. Alternatively, theoretical analyses and numerical simulations can predict soil-pile interactions with low cost, low risk and high efficiency. They also offer inherent mathematical relationships among influencing factors with strong interpretability, and their applicability can be extended by modifying input parameters (Marshall and Haji 2015; Franza et al. 2020; Cao et al. 2021; Li et al. 2024; Lin et al. 2024; Liu et al. 2024). Despite these advantages, both analytical and numerical results are normally derived under simplified conditions, such as idealized geometries, constitutive models and boundary conditions. It can be concluded that the purely data-driven or physics-informed approach exhibits distinct strengths and limitations. Instead of considering them as competing methods, efforts to integrate their advantages are more essential for better understanding and predicting soil-pile interactions induced by tunneling.

Recent advances in physics-informed neural networks (PINNs) have enabled the combination of data and physical laws within neural network frameworks (Raissi et al. 2019; Cai et al. 2021; Karniadakis et al. 2021; Wang and Perdikaris 2021; Song et al. 2024). This is achieved by incorporating experimental data and physical law into a total loss function and then minimizing the loss by gradient-descent methods. Nowadays PINNs have been widely applied to various geotechnical engineering problems, such as consolidation (Lu and Mei 2022; Vahab et al. 2023a; Lan et al. 2024; Yuan et al. 2024; Zhang et al. 2024), tunnelling (Wang et al. 2024; Elbaz et al. 2025; Shen et al. 2025), unsaturated soil mechanics (Yang et al. 2025a; Zhou et al. 2025), cavity expansion analyses (Chen et al. 2024; Yang et al. 2025b), and soil-structure interaction (Cai et al. 2025; Taraghi et al. 2025). In particular, a few studies have also applied PINNs to soil-pile interaction analyses (Madianos et al. 2023; Vahab et al. 2023b; Ouyang et al. 2024). While the applications of PINN have shown great success, a major limitation is their high computational cost, as the network training requires long time. In practice the training efficiency should be as high as possible, especially for engineering problems requiring real-time monitoring and early warning (i.e. continuously update measured data and train the networks).

To fill the gap, this paper innovates a data-driven and physics-informed framework for analyzing tunnelling-induced soil-pile interactions using a rapid variant of PINN, namely the physics-informed extreme learning machine (PIELM) (Dwivedi and Srinivasan (2020)). In this approach, experimental data and governing equations of pile deformation are embedded into a single-layer extreme learning machine (ELM) network. The rest of the paper is organized as follows. Section 2 presents the problem definition and assumptions. Section 3 details the PIELM framework developed for tunneling-induced soil-pile interactions, together with a FDM as the benchmark solution. Validation studies, parametric investigations and illustrative applications are shown in Sections 4, 5 and 6, respectively. Section 7 concludes the paper with key findings.

## 2. Problem Definition and Assumptions

The analysis of soil-pile interaction induced by tunnel excavation is simplified into the mechanical model shown in Figure 1. A single vertical pile with diameter $D$ and length $L$ is considered, adjacent to a tunnel of radius $R$ to be excavated. The vertical distance from the tunnel center to the ground surface is denoted as $H$, and the horizontal distance from the tunnel center to the pile axis is represented by $x_0$. Tunnel excavation disturbs the surrounding soil and induces additional internal forces and deformation in the pile. Following the widely adopted two-stage method (Huang et al. 2009; Mu et al. 2012; Zhang et al. 2018), the tunnelling-induced soil-pile interaction is decomposed into: (i) tunnelling-induced ground deformation (ii) and soil-pile interaction subjected to this ground deformation. Both the pile and soil are assumed to be homogeneous and isotropic. The pile is simplified as the linear-elastic Euler-Bernoulli beam (Zhang et al. 2018; Cao et al. 2021), and the surrounding soil is modelled as a Pasternak foundation (Lin et al. 2020; Liang et al. 2021), with full contact between pile and soil.

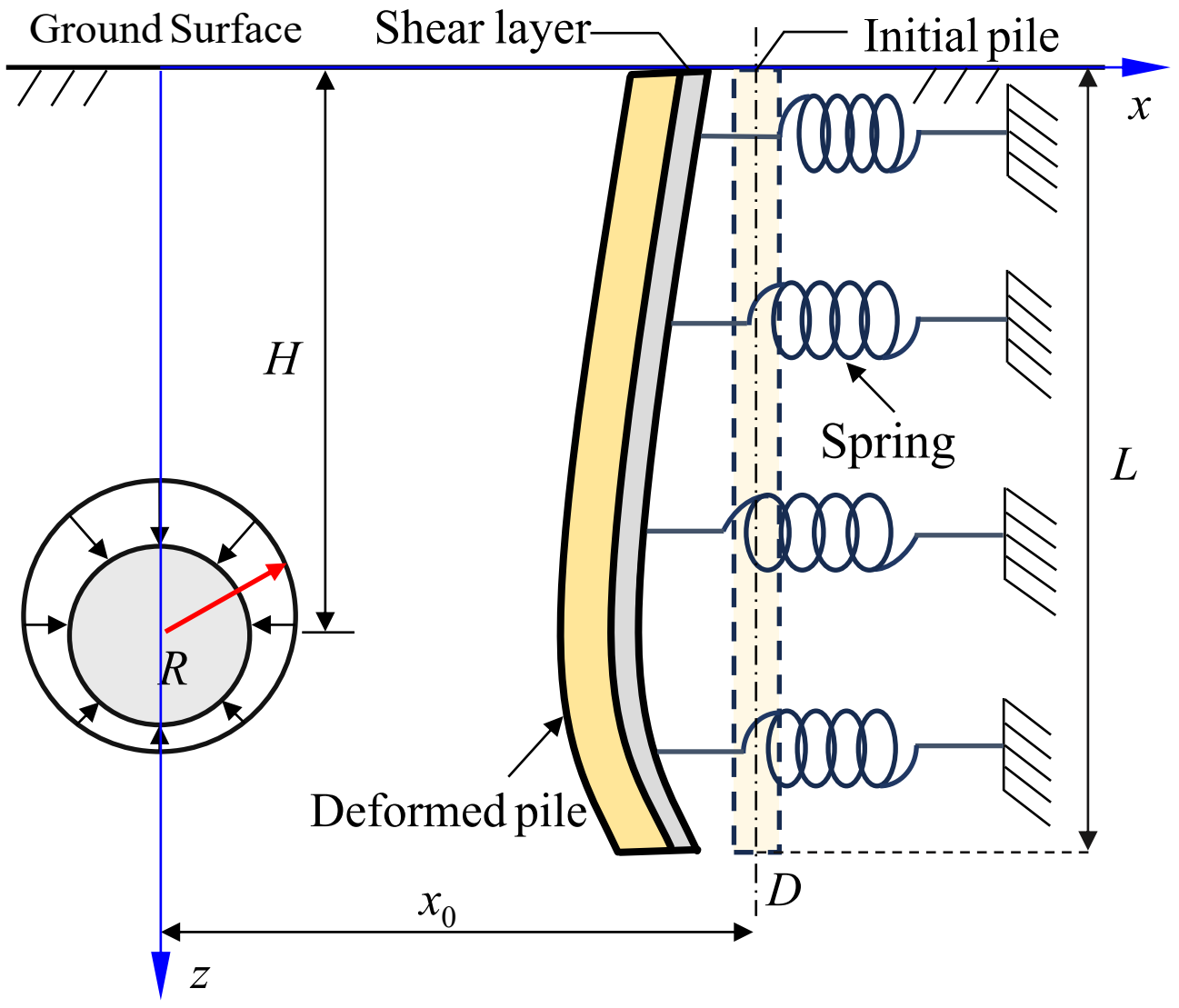

**Figure 1 Schematic of tunnelling-induced soil-pile interaction**

With the above assumptions, the ordinary differential equation (ODE) that governs pile deflection can be expressed as (Cao et al. 2021)

$$EI\frac{\mathrm{d}^4 w(z)}{\mathrm{d}z^4} - GD\frac{\mathrm{d}^2 w(z)}{\mathrm{d}z^2} + k(z)Dw(z) = Df(z) \tag{1}$$

$$f(z) = k(z)u(z) - G\frac{\mathrm{d}^2 u(z)}{\mathrm{d}z^2} \tag{2}$$

where $E$ is the elastic modulus of the pile; $I = \pi D^4/64$ is the rotational inertia of the pile; $w(z)$ denotes the lateral deflection of the pile at the depth $z$; $k(z)$ is depth-dependent subgrade reaction modulus; $G$ is the modulus of the shear layer in the Pasternak foundation model; $f(z)$ accounts for the external load acting on the pile; $u$(z) is the lateral soil displacement caused by tunnel excavation. Following Yu et al. (2013) and Tanahashi (2004), $k(z)$ and $G$ are respectively defined as

$$k(z) = \frac{3.08}{\eta}\frac{E_s}{1-\nu_s^2}\left(\frac{E_s D^4}{EI}\right)^{\frac{1}{8}} \tag{3}$$

$$\eta = \begin{cases} 2.18 & z/D \le 0.5 \\ 1 + \dfrac{1}{1.7z/D} & z/D > 0.5 \end{cases} \tag{4}$$

$$G = \frac{E_s t}{6(1+\nu_s)} \tag{5}$$

where $E_s$ and $\nu_s$ are the elastic modulus and Poisson's ratio of the soil, respectively; $t$ is the shear layer thickness with a typical value of 11$D$ (Yu et al. 2013). To describe the lateral soil displacement induced by tunnelling, the analytical expression proposed by Loganathan (1998) is adopted in this paper:

$$u(z) = -\varepsilon R^2 x \left\{ \begin{array}{l} \dfrac{1}{x^2+(H-z)^2} + \dfrac{3-4\nu_s}{x^2+(H+z)^2} \\ -\dfrac{4z(z+H)}{\left[\left(x^2+(H+z)^2\right)\right]^2} \end{array} \right\} \exp\left\{-\left[\frac{1.38x^2}{(H+R)^2} + \frac{0.69z^2}{H^2}\right]\right\} \tag{6}$$

where $\varepsilon$ is the ground volume loss of tunnelling. It should be noted that these formulations inevitably involve simplifying assumptions. In this paper they are adopted primarily as illustrative inputs of physical laws to demonstrate the applicability of PIELM to tunnelling-induced soil-pile interaction analysis.

The boundary conditions for ODE (1) depend on the constraints at the top and tip of the pile, and three typical boundary conditions are considered:

(i) When both the pile top and tip are free, the bending moment $M$ and shear force $Q$ remain zero at the $z$=0 and $z$=$L$:

$$\begin{cases} M(0) = EI\dfrac{\mathrm{d}^2 w}{\mathrm{d}z^2}\bigg|_{z=0} = 0, \ Q(0) = EI\dfrac{\mathrm{d}^3 w}{\mathrm{d}z^3}\bigg|_{z=0} = 0 \\ M(L) = EI\dfrac{\mathrm{d}^2 w}{\mathrm{d}z^2}\bigg|_{z=L} = 0, \ Q(L) = EI\dfrac{\mathrm{d}^3 w}{\mathrm{d}z^3}\bigg|_{z=L} = 0 \end{cases} \tag{7}$$

(ii) When both the top and tip are fixed, the boundary conditions are

$$\begin{cases} w(0) = 0, \ \theta(0) = \dfrac{\mathrm{d}w}{\mathrm{d}z}\bigg|_{z=0} = 0 \\ w(L) = 0, \ \theta(L) = \dfrac{\mathrm{d}w}{\mathrm{d}z}\bigg|_{z=L} = 0 \end{cases} \tag{8}$$

(iii) When the pile top is free and the pile tip is fixed, the boundary conditions are

$$\begin{cases} M(0) = EI\dfrac{\mathrm{d}^2 w}{\mathrm{d}z^2}\bigg|_{z=0} = 0, \ Q(0) = EI\dfrac{\mathrm{d}^3 w}{\mathrm{d}z^3}\bigg|_{z=0} = 0 \\ w(L) = 0\ , \qquad \theta(L) = \dfrac{\mathrm{d}w}{\mathrm{d}z}\bigg|_{z=L} = 0 \end{cases} \tag{9}$$

Now the governing equations for tunnelling-induced soil-pile interactions are established and will be incorporated as the physics-informed component within the PIELM framework in the next section.

## 3. Solution methods

In this section the PIELM framework is detailed with the incorporation of measured data and physical laws.

### *3.1. PIELM framework*

The low training efficiency of PINNs could be attributed to the deep neural network architectures and reliance on gradient-descent methods. A powerful alternative is the PIELM, which replaces the deep neural network with an ELM network (Dwivedi and Srinivasan 2020). Compared to conventional PINNs, there are several modifications in PIELM:

(i) ELM is a single-layer feed-forward network, but there are several hidden layers in deep neural networks;

(ii) The input layer weights in PIELM are randomly generated, while they need to be iteratively trained in PINNs;

(iii) The loss in PIELM is shown in the form of a loss vector, instead of a total loss in PINNs;

(iv) The output weights in PIELM are directly obtained by optimizing the loss vector, abandoning the inefficient gradient-descent methods.

The high accuracy and efficiency of PIELM have been demonstrated by previous studies in solving a wide range of differential equations (Mortari et al. 2019; Dwivedi and Srinivasan 2020; Schiassi et al. 2021; Liu et al. 2023; Ren et al. 2025; Yang et al. 2025c). For example, Schiassi et al. (2021) demonstrated that hard-constrained PIELM is capable of solving both linear and nonlinear ODEs within milliseconds. Ren et al. (2025) employed iterative PIELM to solve Stefan problems, achieving over 98% reduction in training time while improving solution accuracy from $10^{-3}$~$10^{-5}$ to $10^{-6}$~$10^{-8}$. These results highlight the superior efficiency and accuracy of PIELM, making it particularly suitable for analyzing tunnelling-induced soil-pile interaction with measured data.

The PIELM framework for tunnelling-induced soil-pile interaction analysis is illustrated in Figure 2. First, the depth (i.e. $z$ coordinate) is normalized by the pile length $L$, and $N_c$ collocation points $\{\tilde{z}_i\}_{i=1}^{N_c}$ are uniformly generated within the interval [0, 1]. These points are fed into the ELM network with $M_c$ hidden-layer neurons ($\sigma$ is the activation function), and the input weights are randomly generated without training. The ELM network produces four outputs, $g_1$, $g_2$, $g_3$ and $g_4$. The outputs and their derivatives with respect to $\tilde{z}$ are expressed as follows:

$$g_h(\tilde{z};\boldsymbol{\beta}_h)=\sum_{i=1}^{M_c}\beta_{h,i}\sigma(W_i\tilde{z}+b_i)=\sigma(\boldsymbol{W}\tilde{z}+\boldsymbol{b})\boldsymbol{\beta}_h \quad h=1,2,3,4 \tag{10}$$

$$\frac{\mathrm{d}g_h(\tilde{z};\boldsymbol{\beta}_h)}{\mathrm{d}\tilde{z}}=\sum_{i=1}^{M_c}\beta_{h,i}W_i\frac{\mathrm{d}\sigma(W_i\tilde{z}+b_i)}{\mathrm{d}(W_i\tilde{z}+b_i)} \quad h=1,2,3,4 \tag{11}$$

where $\boldsymbol{W}=[w_1,w_2,\dots,w_{M_c}]$ is the input weight vector and $\boldsymbol{b}=[b_1,b_2,\dots,b_{M_c}]$ is the bias vector; $\boldsymbol{\beta}_h=[\beta_{h,1},\beta_{h,2},\dots,\beta_{h,M_c}]^T$ is the output weight vector for the $h$-th output, and the total output weight vector is expressed as $\boldsymbol{\beta}=[\boldsymbol{\beta}_1^T,\boldsymbol{\beta}_2^T,\boldsymbol{\beta}_3^T,\boldsymbol{\beta}_4^T]^T$.

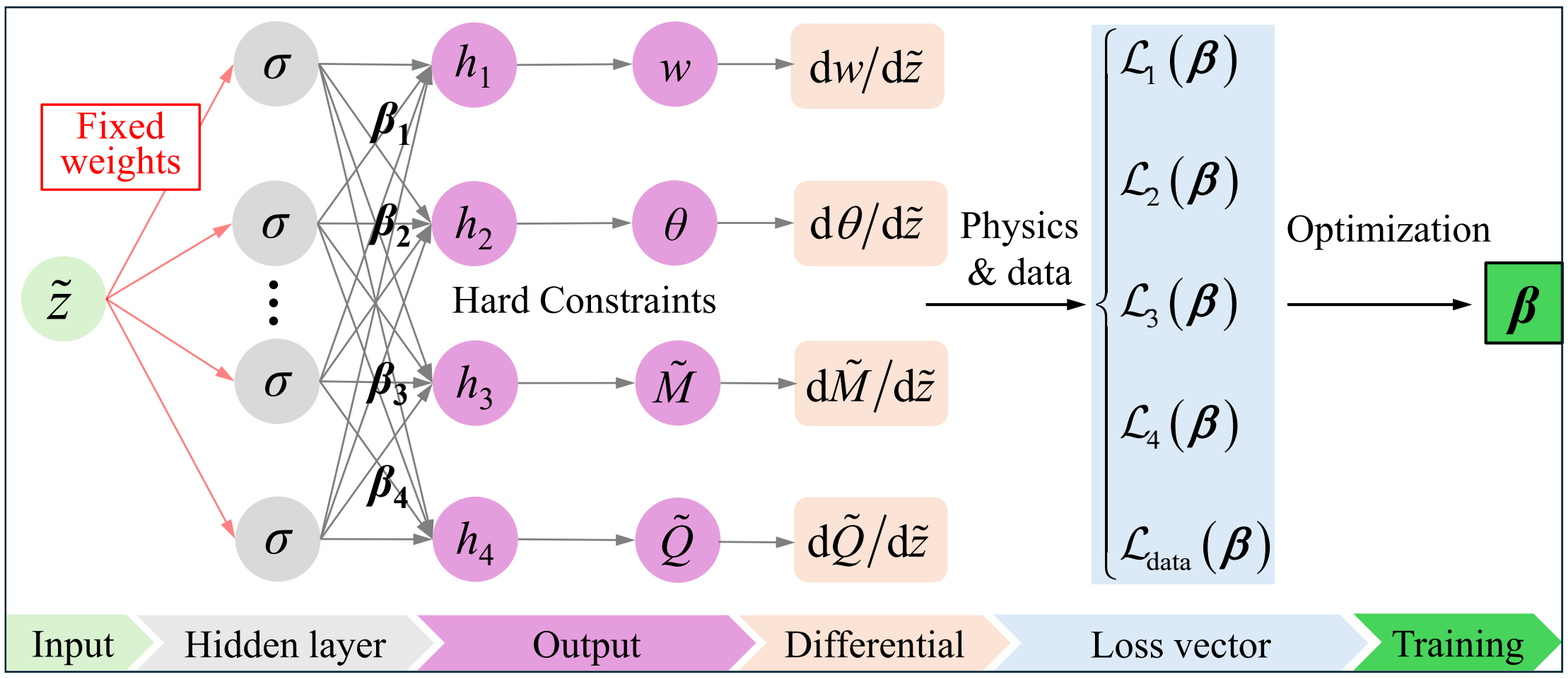

**Figure 2 PIELM framework for soil-pile interactions**

Then, a method for applying hard constraints (Schiassi et al. 2021; Lan et al. 2024) is introduced to improve solution accuracy. Decomposing the fourth-order ODE Eq. (1) into a system of four first-order ODEs, we can get

$$\frac{\mathrm{d}w(z)}{\mathrm{d}z}=\theta(z) \tag{12}$$

$$\frac{\mathrm{d}\theta(z)}{\mathrm{d}z}=\bar{M}(z) \tag{13}$$

$$\frac{\mathrm{d}\bar{M}(z)}{\mathrm{d}z}=\bar{Q}(z) \tag{14}$$

$$EI\frac{\mathrm{d}\bar{Q}(z)}{\mathrm{d}z}-GD\bar{M}(z)+k(z)Dw(z)=Df(z) \tag{15}$$

where $\bar{M}(z)=M/EI$ and $\bar{Q}(z)=Q/EI$ are the normalized bending moment and shear force, respectively. In this way the hard constraints can be easily applied to the PIELM framework with various boundary conditions (see Eqs. (7)~(9)), which makes the constructed solution functions automatically satisfy boundary conditions. Concretely, when both the pile top and tip are free (see Eq. (7)), the solution for ODEs (12)~(15) can be expressed as

$$w(\tilde{z};\boldsymbol{\beta}_1)=h_1(\tilde{z};\boldsymbol{\beta}_1) \tag{16}$$

$$\theta(\tilde{z};\boldsymbol{\beta}_2)=h_2(\tilde{z};\boldsymbol{\beta}_2) \tag{17}$$

$$\bar{M}(\tilde{z};\boldsymbol{\beta}_3)=h_3(\tilde{z};\boldsymbol{\beta}_3)+(\tilde{z}-1)h_3(0;\boldsymbol{\beta}_3)-\tilde{z}h_3(1;\boldsymbol{\beta}_3) \tag{18}$$

$$\bar{Q}(\tilde{z};\boldsymbol{\beta}_4)=h_4(\tilde{z};\boldsymbol{\beta}_4)+(\tilde{z}-1)h_4(0;\boldsymbol{\beta}_4)-\tilde{z}h_4(1;\boldsymbol{\beta}_4) \tag{19}$$

When both the top and tip are fixed (see Eq. (8)), the expressions of solutions are defined as

$$w(\tilde{z};\boldsymbol{\beta}_1)=h_1(\tilde{z};\boldsymbol{\beta}_1)+(\tilde{z}-1)h_1(0;\boldsymbol{\beta}_1)-\tilde{z}h_1(1;\boldsymbol{\beta}_1) \tag{20}$$

$$\theta(\tilde{z};\boldsymbol{\beta}_2)=h_2(\tilde{z};\boldsymbol{\beta}_2)+(\tilde{z}-1)h_2(0;\boldsymbol{\beta}_2)-\tilde{z}h_2(1;\boldsymbol{\beta}_2) \tag{21}$$

$$\bar{M}(\tilde{z};\boldsymbol{\beta}_3)=h_3(\tilde{z};\boldsymbol{\beta}_3) \tag{22}$$

$$\bar{Q}(\tilde{z};\boldsymbol{\beta}_4)=h_4(\tilde{z};\boldsymbol{\beta}_4) \tag{23}$$

When the pile top is free and the pile tip is fixed (see Eq. (9)), the expressions are given by

$$w(\tilde{z};\boldsymbol{\beta}_1)=h_1(\tilde{z};\boldsymbol{\beta}_1)-h_1(1;\boldsymbol{\beta}_1) \tag{24}$$

$$\theta(\tilde{z};\boldsymbol{\beta}_2)=h_2(\tilde{z};\boldsymbol{\beta}_2)-h_2(1;\boldsymbol{\beta}_2) \tag{25}$$

$$\bar{M}(\tilde{z};\boldsymbol{\beta}_3)=h_3(\tilde{z};\boldsymbol{\beta}_3)-h_3(0;\boldsymbol{\beta}_3) \tag{26}$$

$$\bar{Q}(\tilde{z};\boldsymbol{\beta}_4)=h_4(\tilde{z};\boldsymbol{\beta}_4)-h_4(0;\boldsymbol{\beta}_4) \tag{27}$$

Later, after outputting the solution expressions and their derivatives, a loss vector $\mathcal{L}$ can be derived by combining the physical laws and measured data as

$$\boldsymbol{\mathcal{L}}=\left[\left\{\mathcal{L}_1(\boldsymbol{\beta}_1)_i\right\}_{i=1}^{N_c},\left\{\mathcal{L}_2(\boldsymbol{\beta}_2)_i\right\}_{i=1}^{N_c},\left\{\mathcal{L}_3(\boldsymbol{\beta}_3)_i\right\}_{i=1}^{N_c},\left\{\mathcal{L}_4(\boldsymbol{\beta}_4)_i\right\}_{i=1}^{N_c},\left\{\mathcal{L}_{\text{data}}(\boldsymbol{\beta}_1)_j\right\}_{j=1}^{N_\text{d}}\right]\to\mathbf{0} \tag{28}$$

$$\mathcal{L}_1(\boldsymbol{\beta})_i=\frac{1}{L}\frac{\mathrm{d}w(\tilde{z}_i;\boldsymbol{\beta}_1)}{\mathrm{d}\tilde{z}}-\theta(\tilde{z}_i;\boldsymbol{\beta}_2) \tag{29}$$

$$\mathcal{L}_2(\boldsymbol{\beta})_i=\frac{1}{L}\frac{\mathrm{d}\theta(\tilde{z}_i;\boldsymbol{\beta}_2)}{\mathrm{d}\tilde{z}}-\bar{M}(\tilde{z}_i;\boldsymbol{\beta}_3) \tag{30}$$

$$\mathcal{L}_3(\boldsymbol{\beta})_i=\frac{1}{L}\frac{\bar{M}(\tilde{z}_i;\boldsymbol{\beta}_3)}{\mathrm{d}\tilde{z}}-\bar{Q}(\tilde{z}_i;\boldsymbol{\beta}_4) \tag{31}$$

$$\mathcal{L}_4(\boldsymbol{\beta})_i=\frac{EI}{L}\frac{\mathrm{d}\bar{Q}(\tilde{z}_i;\boldsymbol{\beta}_4)}{\mathrm{d}\tilde{z}}-GD\bar{M}(\tilde{z}_i;\boldsymbol{\beta}_3)+k(Lz_i)Dw(\tilde{z}_i;\boldsymbol{\beta}_1)-Df(Lz_i) \tag{32}$$

$$\mathcal{L}_{\text{data}}(\boldsymbol{\beta})_j=w(\tilde{z}_j;\boldsymbol{\beta}_1)-w_{\text{data}}(\tilde{z}_j) \tag{33}$$

where $\boldsymbol{\mathcal{L}}_1(\boldsymbol{\beta})$, $\boldsymbol{\mathcal{L}}_2(\boldsymbol{\beta})$, $\boldsymbol{\mathcal{L}}_3(\boldsymbol{\beta})$ and $\boldsymbol{\mathcal{L}}_4(\boldsymbol{\beta})$ are the loss vector components corresponding to Eqs. (12), (13), (14) and (15), respectively; $\boldsymbol{\mathcal{L}}_{\text{data}}(\boldsymbol{\beta})$ denotes the component associated with the monitored data. It is also necessary to note that the loss vector components involved by boundary conditions vanish in Eq. (28) as the hard constraints are applied.

Finally, the ELM network will be trained by directly calculating $\boldsymbol{\beta}$ using the least-squares method with the Moore-Penrose generalized inverse. The lateral deformation, the rotary angle, bending moment and shear force can also be readily obtained with the known $\boldsymbol{\beta}$:

$$w(z)\ =w\left(\frac{z}{L};\boldsymbol{\beta}_1\right) \tag{34}$$

$$\theta(z)\ =\theta\left(\frac{z}{L};\boldsymbol{\beta}_2\right) \tag{35}$$

$$M(z)\ =EI\cdot\bar{M}\left(\frac{z}{L};\boldsymbol{\beta}_3\right) \tag{36}$$

$$Q(z)\ =EI\cdot\bar{Q}\left(\frac{z}{L};\boldsymbol{\beta}_4\right) \tag{37}$$

### *3.2. FDM benchmark solution*

As the additional external load caused by tunneling is highly complex, an analytical solution to ODE (1) is not easily available, if not impossible. Therefore, a numerical solution using the finite difference method (FDM) is adopted as a baseline for validating the PIELM approach.

As shown in Figure 3, the pile is discretized into $N_f$ segments with $N_f$ +1 nodes and each length of $l = L/N_f$ . Four additional virtual nodes are added at the tip and top of the pile to deal with boundary conditions: $j$=-2, -1, $N_f$, and $N_f$+1, where $j$ is the node number. The central difference method is used to compute the derivatives of pile deflection, as

$$w_j^{(1)} = \frac{\mathrm{d}^2 w(z)}{\mathrm{d}z^2} = \frac{w_{j+1} - w_{j-1}}{2l} \tag{38}$$

$$w_j^{(2)} = \frac{\mathrm{d}^2 w(z)}{\mathrm{d}z^2} = \frac{w_{j+1} - 2w_j + w_{j-1}}{l^2} \tag{39}$$

$$w_j^{(3)} = \frac{\mathrm{d}^3 w(z)}{\mathrm{d}z^3} = \frac{w_{j+2} - 2w_{j+1} + 2w_{j-1} - w_{j-2}}{2l^3} \tag{40}$$

$$w_j^{(4)} = \frac{\mathrm{d}^4 w(z)}{\mathrm{d}z^4} = \frac{w_{j+2} - 4w_{j+1} + 6w_j - 4w_{j-1} + w_{j-2}}{l^4} \tag{41}$$

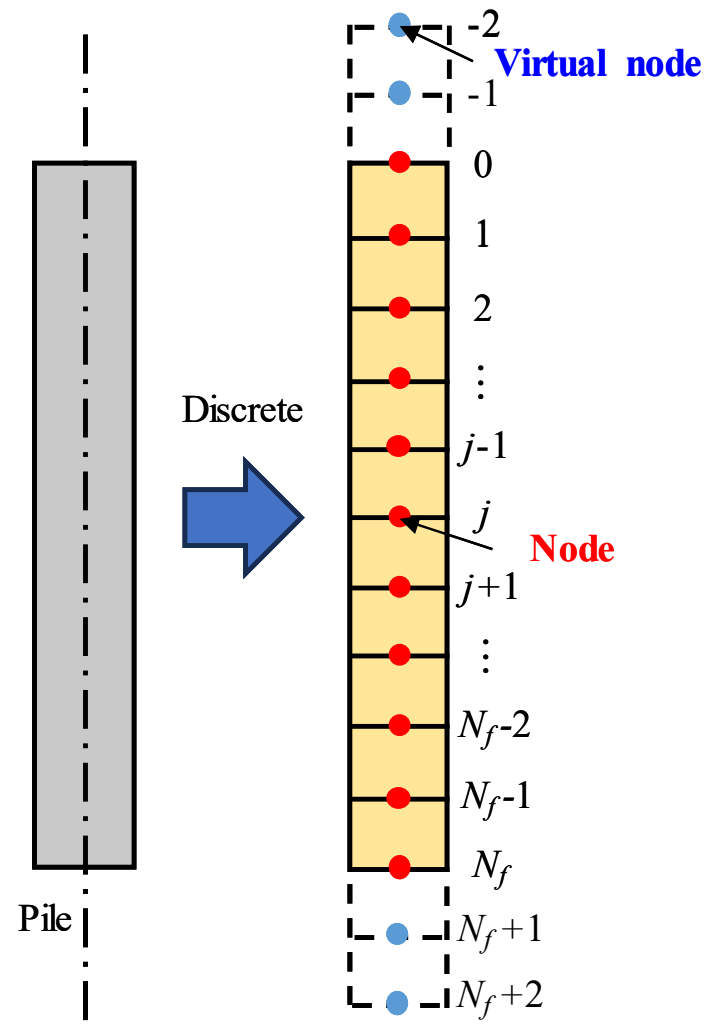

**Figure 3 Discretization of pile for FDM**

Substituting Eqs. (39) and (41) into Eq. (1), the ODE (1) for pile deflection can be transformed as

$$\frac{EI}{D}\frac{w_{j+2}-4w_{j+1}+6w_j-4w_{j-1}+w_{j-2}}{l^4}-G\frac{w_{j+1}-2w_j+w_{j-1}}{l^2}+k\left(z_j\right)w_j=f\left(z_j\right) \quad (42)$$

The set of discrete equations for all interior nodes can be assembled into a global system of linear equations, which is shown in the matrix as:

$$\mathbf{K}\left\{w_j\right\}_{j=1}^{N_f}=\left\{f_j\right\}_{j=1}^{N_f} \quad (43)$$

where $f_j$ is the external load at the selected nodes, determined by Eq. (2); **K** is the global stiffness matrix and is detailed in the Appendix for various boundary conditions. Upon solving Eq. (43), the rotation angle $\theta$, bending moment $M$ and shear force $Q$ can be calculated by

$$\theta_j=\frac{1}{2l}\left(w_{j+1}-w_{j-1}\right) \quad (44)$$

$$M_j=\frac{EI}{l^2}\left(w_{j+1}-2w_j+w_{j-1}\right) \quad (45)$$

$$Q_j=\frac{EI}{2l^3}\left(w_{j+2}-2w_{j+1}+2w_{j-1}-w_{j-2}\right) \quad (46)$$

## 4. Validation of the PIELM approach

The PIELM method for tunnelling-induced soil-pile interaction analysis is first validated by comparison with results obtained from the boundary element method BEM and FDM. The input parameters used in this section are summarized in Table 1.

At first, the PIELM approach is validated against the BEM solution of Xu and Poulos (2001). Figure 4 presents the comparison of lateral displacement and bending moments between the PIELM predictions and BEM calculations. The pile top and tip are free, and the used input parameters are listed in Table 1 with three different levels of ground volume loss. The results show that PIELM accurately captures the lateral deflection and bending moment of the pile, showing good agreement with the BEM results and thereby validating the accuracy of the PIELM approach.

**Table 1 Input parameters for physics (after Xu and Poulos (2001)) and network**

| Parameters | $E$(MPa) | $v$ | $E_s$(GPa) | $D$(m) | $L$(m) | $H$(m) | $R$(m) | $x_0$(m) |
|---|---|---|---|---|---|---|---|---|
| Value | 24 | 0.5 | 30 | 0.5 | 25 | 20 | 3 | 4.5 |
| Parameters | $M_c$ | $N_c$ | $N_f$ | $N_{data}$ | $\sigma$ | Computer | | |
| Value | 500 | 1000 | 2000 | 0 | Tanh | i9 14900 16G RAM | | |

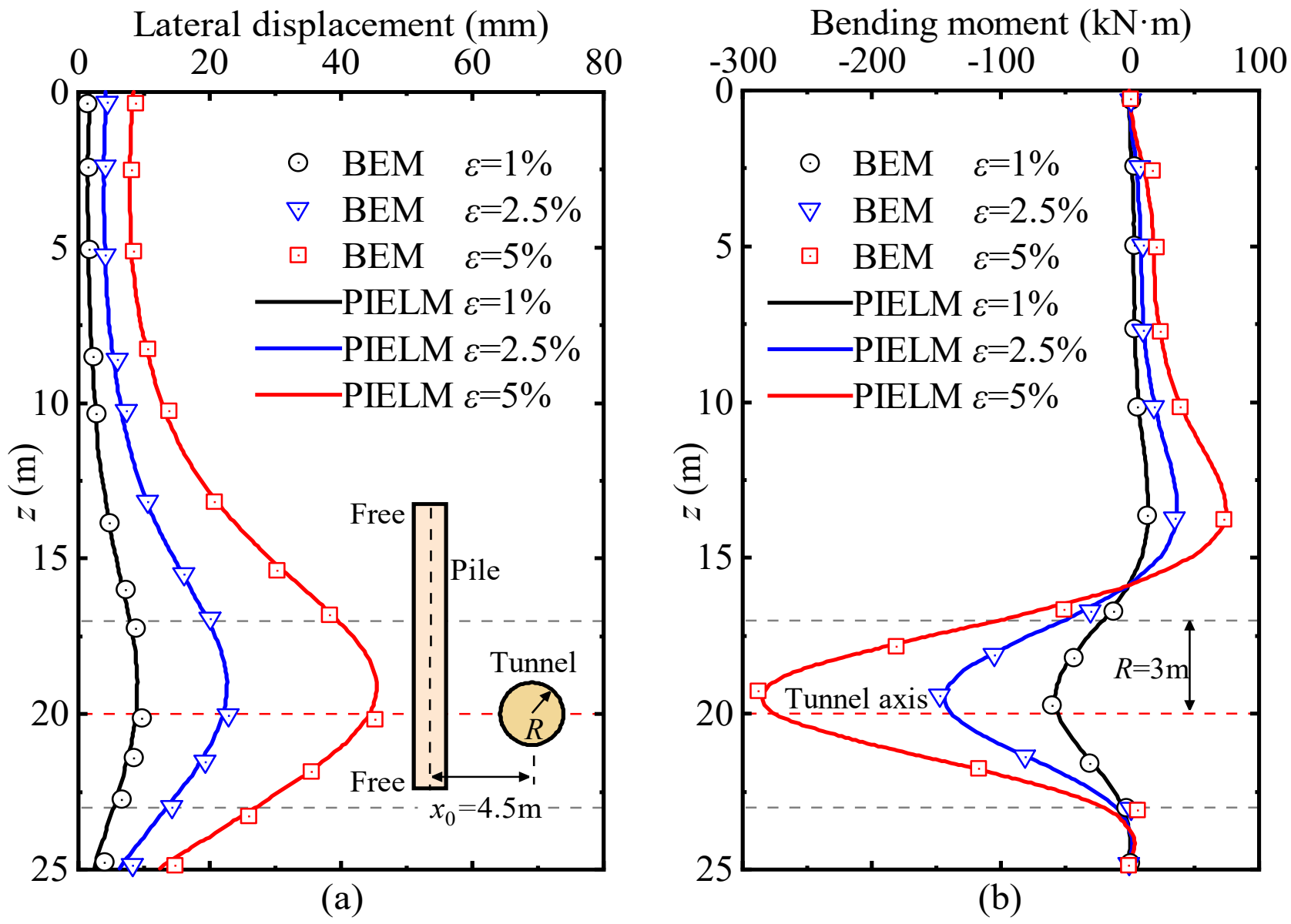

**Figure 4 Comparison of pile lateral deflections and bending moments between PIELM and boundary element method**

The PIELM approach is further verified by comparison with the FDM solution. Three cases with different boundary conditions are selected for validation: (i) the pile is free at the pile top and tip; (ii) the pile is fixed at the pile top and tip; and (iii) the pile top is free while the pile tip is fixed. Figure 5 presents the comparisons of lateral deflection, bending moment, and shear force for the piles in all three cases. The mechanical response of the piles is significantly influenced by tunneling-induced volume loss, and the effects of boundary conditions are primarily observed within a certain range near the pile tip/top. Overall, PIELM predictions closely match the numerical results obtained by FDM, and hence the power of the PIELM approach is verified again.

To quantify the accuracy more rigorously, the relative $L_2$ error is adopted, defined by

$$L_2 = \frac{\sqrt{\sum_{1}^{N_{\text{test}}} (\text{FDM-PIELM})^2}}{\sqrt{\sum_{1}^{N_{\text{test}}} (\text{FDM})^2}} \tag{47}$$

where $N_{\text{test}}$ is the number of test points. Table 2 summarizes the relative $L_2$ errors of lateral displacements, bending moments, and shear forces in these cases. We can find that $L_2$ for lateral pile displacement, bending moment and shear force are on the order of $10^{-6}$~$10^{-3}$, $10^{-5}$~$10^{-2}$ and $10^{-3}$~$10^{-2}$, respectively. The maximum $L_2$ is less than 1.4% and the training time is less than 1s, which are accurate and efficient enough for real-time soil-pile interaction analysis.

**Table 2** **Relative $L_2$ errors of PIELM**

| Boundary conditions | Ground volume loss | $L_2$ error for $w$ | $L_2$ error for $M$ | $L_2$ error for $Q$ | Training time (s) |
|---|---|---|---|---|---|
| Free top and tip | $\varepsilon$=1% | 1.23e-06 | 4.35 e-05 | 5.11e-03 | 0.593 |
| | $\varepsilon$=2% | 1.99e-06 | 4.33 e-05 | 5.11e-03 | 0.610 |
| | $\varepsilon$=3% | 1.75e-06 | 4.32 e-05 | 5.11e-03 | 0.584 |
| Fixed top and free tip | $\varepsilon$=1% | 1.41 e-03 | 1.3 e-02 | 1.14 e-02 | 0.586 |
| | $\varepsilon$=2% | 1.41 e-03 | 1.3 e-02 | 1.14 e-02 | 0.641 |
| | $\varepsilon$=3% | 1.41 e-03 | 1.3 e-02 | 1.14 e-02 | 0.584 |
| Fixed top and tip | $\varepsilon$=1% | 1.47 e-03 | 1.32 e-02 | 1.12 e-02 | 0.630 |
| | $\varepsilon$=2% | 1.47 e-03 | 1.32 e-02 | 1.12 e-02 | 0.555 |
| | $\varepsilon$=3% | 1.47 e-03 | 1.32 e-02 | 1.12 e-02 | 0.592 |

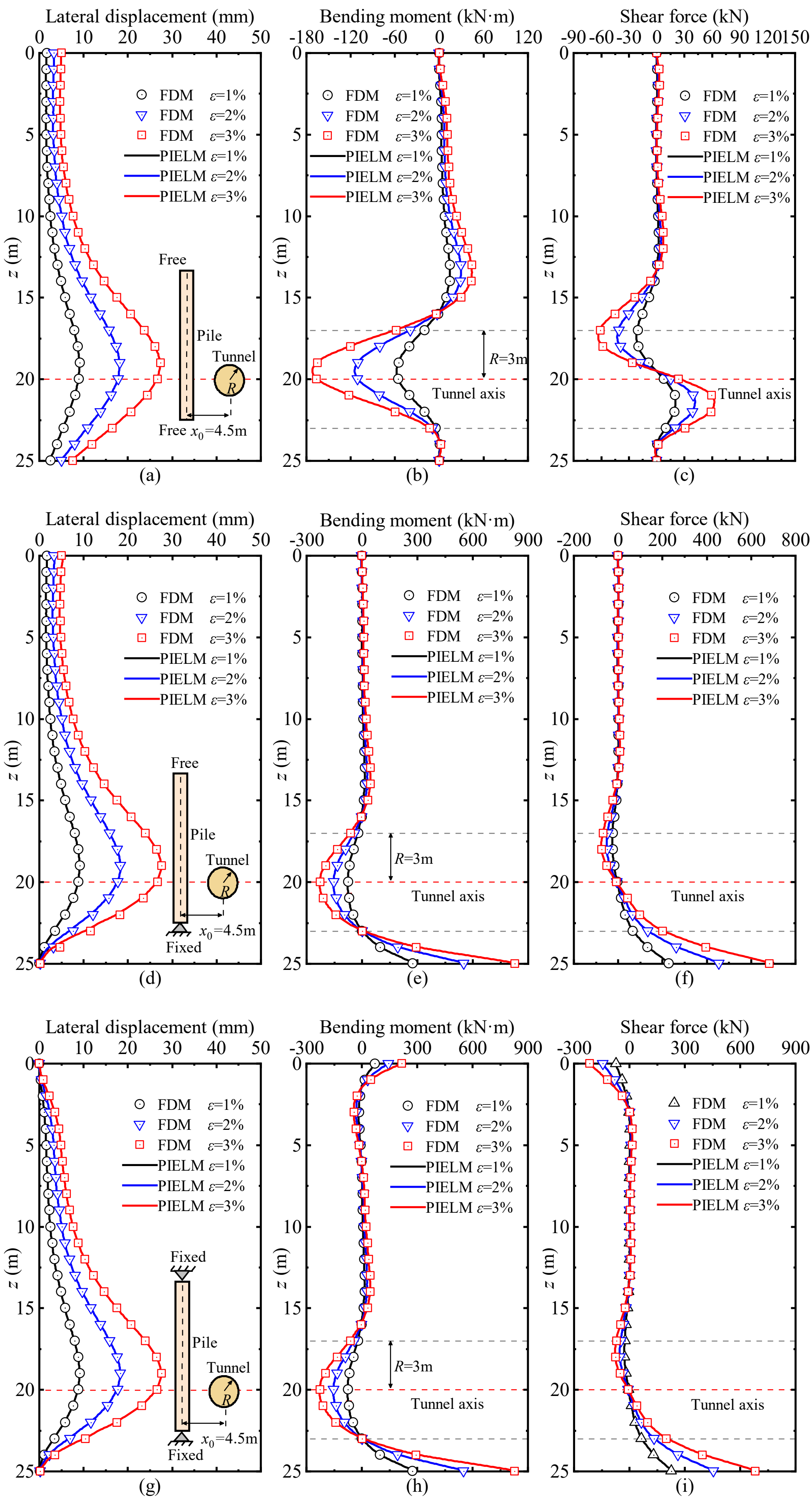

**Figure 5 Comparison of pile lateral displacements, bending moments, and shear forces between the PIELM and FDM**

## 5. Parametric studies

This section shows a series of parametric studies to investigate the performance of PIELM, considering the ELM architecture, data monitoring locations, and data monitoring numbers. In the following the pile is assumed to be free at pile top and tip, and the ground volume loss $\varepsilon$ is set to 1%. Unless stated otherwise, all other parameters are the same as those in Table 1.

### *5.1. ELM architecture*

The PIELM framework is simple yet powerful, with two main hyperparameters: the number of collection points $N_c$ (excluding measured data) and the number of hidden layer neurons $M_c$. The sensitivity of PIELM to $N_c$ and $M_c$ is examined in this subsection, and the results are summarized in Table 3 and Table 4. For a fixed number of training points (e.g., $N_c$=500), the accuracy of PIELM generally increases as $M_c$ increases, but further increasing $M_c$ from 100 to 500 does not significantly reduce the relative $L_2$ error. This is primarily because: (i) overfitting appears when too many neurons are used; and (ii) FDM solution is not rigorously accurate when $L_2$ is too small. The prediction accuracy of PIELM decreases successively for pile deflection, bending moment, and shear force, which is consistent with the increasing derivative order of pile deflection. A similar trend is also observed when increasing training points with fixed neuron number (e.g., $M_c$=50), where $L_2$ is hardly reduced when $N_c$ increases from 400 to 1000. For network training, it takes tens to hundreds of milliseconds according to the numbers of collection points and neurons, showing the high training efficiency of PIELM.

Meanwhile, five repeated calculations are performed to assess the robustness of PIELM, as the hidden-layer weights are randomly assigned prior to training. The relative errors and training time are reported in Table 5. These results confirm that random initialization of input-layer weights has little impact on the performance of PIELM. Overall, the efficiency, accuracy and robustness of the PIELM framework for tunnelling-induced soil-pile interactions have been demonstrated. Once network parameters are properly selected, PIELM provides solutions in the order of 1e-06 for deflection and 1e-05 for bending moments, only requiring millisecond-level time to train the network. This superior performance of PIELM is thereby suitable for real-time monitoring of the mechanical response of piles, as will be discussed next.

**Table 3 PIELM performance with neuron numbers.**

| $M_c$ | $L_2$ error for $w$ | $L_2$ error for $M$ | $L_2$ error for $Q$ | Training time (s) |
|---|---|---|---|---|
| 25 | 6.68 e-04 | 2.12 e-03 | 1.30 e-02 | 0.008 |
| 50 | 6.26e-06 | 7.91 e-05 | 5.12 e-03 | 0.019 |
| 100 | 1.52 e-06 | 4.42 e-05 | 5.11 e-03 | 0.033 |
| 500 | 1.23e-06 | 4.35 e-05 | 5.11e-03 | 0.054 |

Note: the training point number is fixed at 500

**Table 4 PIELM performance with training point numbers**

| $N_c$ | $L_2$ error for $w$ | $L_2$ error for $M$ | $L_2$ error for $Q$ | Training time (s) |
|---|---|---|---|---|
| 25 | 6.93 e-03 | 3.20e-02 | 8.38e-02 | 0.002 |
| 50 | 2.58e-04 | 2.76 e-03 | 5.65 e-03 | 0.002 |
| 100 | 6.67 e-05 | 2.96 e-04 | 5.14 e-03 | 0.006 |
| 400 | 1.45e-06 | 5.82 e-05 | 5.12e-03 | 0.014 |
| 1000 | 1.37e-06 | 4.66 e-05 | 5.12e-03 | 0.027 |

Note: the neural number is fixed at 50

**Table 5 PIELM performance with test numbers.**

| Test number | $L_2$ error for $w$ | $L_2$ error for $M$ | $L_2$ error for $Q$ | Training time (s) |
|---|---|---|---|---|
| 1 | 1.24 e-06 | 4.34 e-05 | 5.11 e-03 | 0.641 |
| 2 | 1.25 e-06 | 4.33 e-05 | 5.11 e-03 | 0.671 |
| 3 | 1.26 e-06 | 4.33 e-05 | 5.11 e-03 | 0.752 |
| 4 | 1.23 e-06 | 4.33 e-05 | 5.11 e-03 | 0.616 |
| 5 | 1.25 e-06 | 4.35 e-05 | 5.11 e-03 | 0.614 |

Note: the neuron number and training points are 500 and 1000, respectively.

### 5.2. *Data monitoring locations*

In the above analysis, only training points were used in the PIELM, while monitored data were excluded. This type of analysis is typically referred to as the forward problem. To emphasize the importance of the data-driven and physics-informed PIELM approach, monitored data of pile deflection are incorporated into the PIELM framework, and the influence of their spatial locations is first examined.

A series of seven case studies on the locations of monitored data is conducted, as summarized in Table 6. The "monitored" data are pseudo-observations generated from the benchmark FDM solution. In Series 1~6 (S1~S6), five monitored data points are selected, with $M_c$=$N_c$=20. S7 serves as a control group, where $M_c$ =20 and $N_c$ =25, ensuring the same total number of training points (i.e., $N_c$+$N_{data}$=25). Smaller $M_c$ and $N_c$ are chosen to magnify the error, making it easier to identify the optimal monitoring locations.

**Table 6　Series of parametric studies with various data monitoring locations**

| Series | Data monitoring locations $z$ (m) | $L_2$ error $w$ | $L_2$ error for $M$ |
|---|---|---|---|
| S1 | 0, 1, 2, 3, 4 | 4.45 e-03 | 4.30 e-02 |
| S2 | 5, 6.5, 8, 9.5, 11 | 3.23 e-03 | 3.40 e-02 |
| S3 | 11, 12.5, 14, 15.5, 17 | 2.72 e-03 | 3.26 e-02 |
| S4 | 17, 18.5, 20, 21.5, 23 | 1.13 e-03 | 2.10 e-02 |
| **S5** | **0, 17, 20, 23, 25** | **7.88 e-04** | **1.13 e-02** |
| S6 | 0, 6.25, 12.5, 18.75, 25 | 3.44 e-03 | 3.55 e-02 |
| S7 | No monitored data but $N_c$=25 | 3.67 e-02 | 1.71 e-01 |

Figure 6 shows the distribution of absolute errors in pile deflection and bending moment for Series 1~7, where monitored data are marked as solid symbols for clarity. In Series 1~4 (Figure 6 (a)), the absolute errors within the monitoring zones are significantly reduced compared to Series 7, and the absolute errors decrease by one to two orders of magnitude. The accuracy improvement strongly depends on the spatial locations of the monitored data. For instance, placing monitored data near the pile top/tip and tunneling zones yields higher accuracy, as these regions exhibit larger gradients of deflection and bending moment that can be effectively constrained by the "exact" data. Moreover, the improvement for deflection predictions is more pronounced than that for bending moment, because the monitored data are provided in terms of pile deflections.

To further demonstrate the significance of data monitoring loactions, in S5 we assign two monitoring points at the pile top/tip and three near the tunneling zone. As expected, S5 achieves the highest accuracy, with relative $L_2$ errors reduced by 77.1% and 68.2% for $w$ and $M$, respectively, compared with S6 (the monitored data are uniformly distributed). These findings indicate that collecting monitored data in critical regions, such as the pile top/tip and tunneling zones, is substantially more effective.

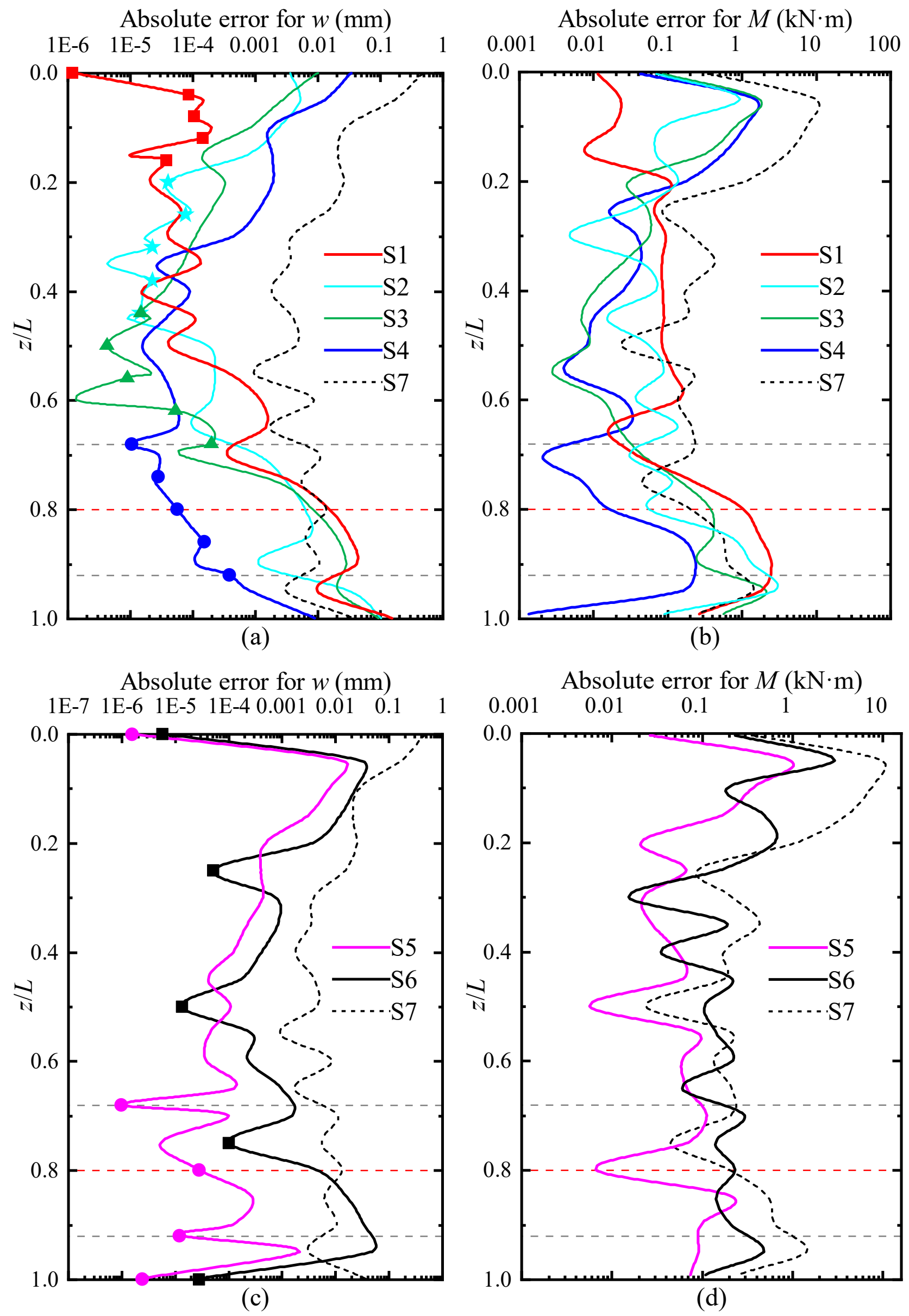

**Figure 6 Distribution of absolute errors with data monitoring locations: (a) lateral displacements; (b) bending moments; (c) lateral displacements; (d) bending moments.**

### *5.3. Data monitoring numbers*

After examining the influence of data monitoring locations, this subsection investigates the effect of data number on the prediction accuracy of PIELM, as shown in Figure 7 and Table 7. In S8~S13 $M_c$=$N_c$=20 are selected, and the number of monitored data increases from 0 to 10 near the tunneling zone. S8 is a control group with zero monitored data. Similarly, the absolute errors in the regions containing monitored data can be significantly reduced. The overall prediction accuracy of PIELM gradually increases with the number of monitored data. For a clearer illustration, the variation of $L_2$ with the data monitoring number is plotted in Figure 8. Compared to S8 (without monitored data), adding 4~5 monitored data can greatly reduce $L_2$ from 5.05% to 0.58% for $w$ and 19.8% to 4.9% for $M$. However, further increase of monitored data from 5 to 10 does not significantly improve the predictive accuracy, as shown in Figure 8. Considering the cost of data collection, the optimal number of monitored data is likely 4 or 5 with the given input parameters. The results suggest that incorporating only a few monitoring points is sufficient to considerably improve the accuracy of PIELM in soil-pile interaction analysis.

**Table 7 Series of parametric studies with locations of monitored data**

| Series | Data monitoring locations $z$ (m) | $L_2$ error for $w$ | $L_2$ error for $M$ |
|---|---|---|---|
| S8 | No monitored data and $N_c$=20 | 5.05 e-02 | 1.98 e-01 |
| S9 | 17, 23 | 1.77 e-02 | 8.87 e-02 |
| S10 | 17, 20, 23 | 5.80 e -03 | 4.89 e -02 |
| S11 | 17, 18.5, 21.5, 23 | 3.43 e -03 | 3.36 e -02 |
| S12 | 17, 18.5, 20, 21.5, 23 | 1.13 e-03 | 2.10 e-02 |
| S13 | 17, 17.5, 18, 18.5, 19, 20, 21, 21.5, 22, 23 | 8.50e-03 | 1.28e-02 |

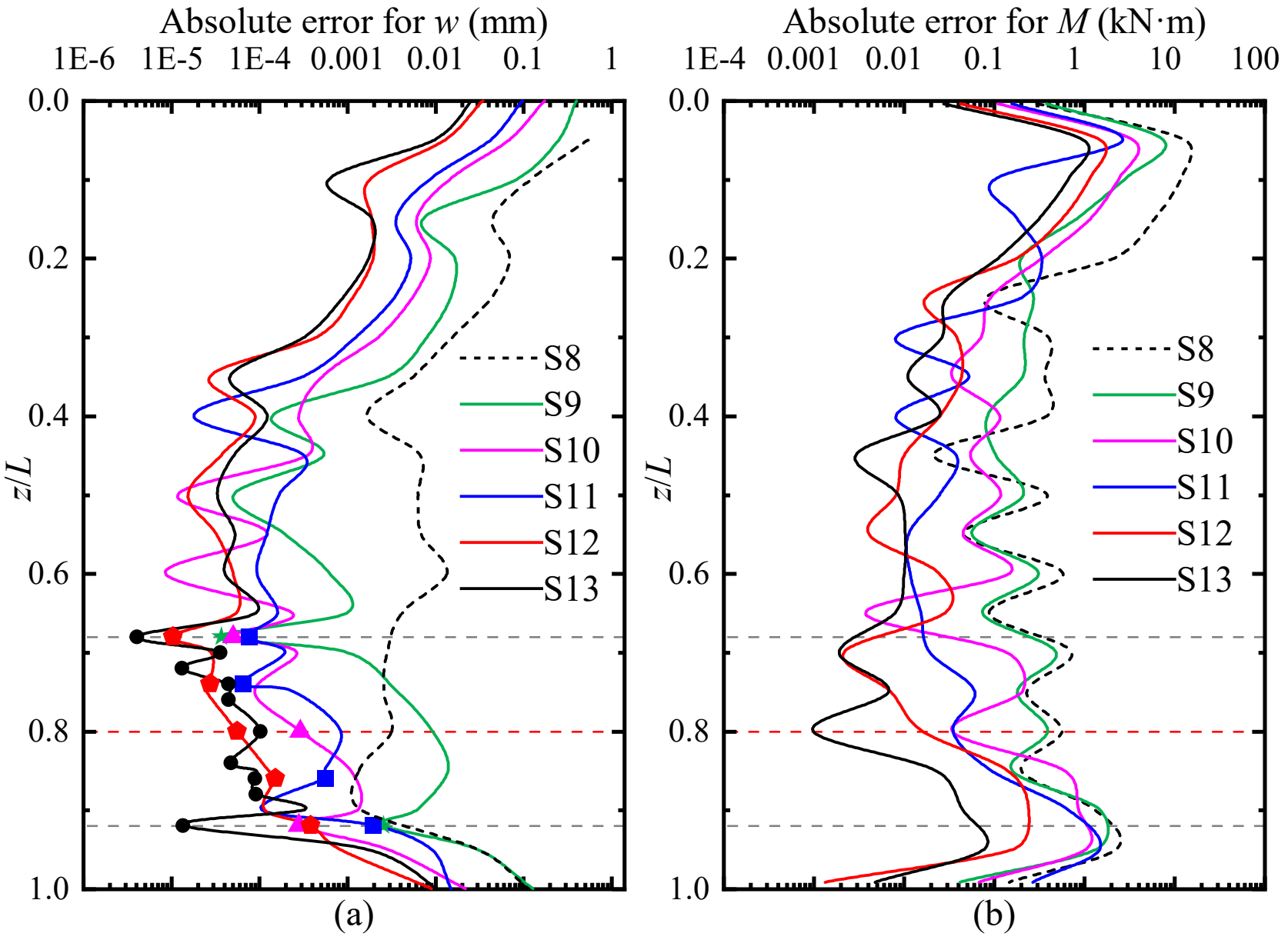

**Figure 7 Distribution of absolute errors with data monitoring numbers: (a) lateral displacements; (b) bending moments.**

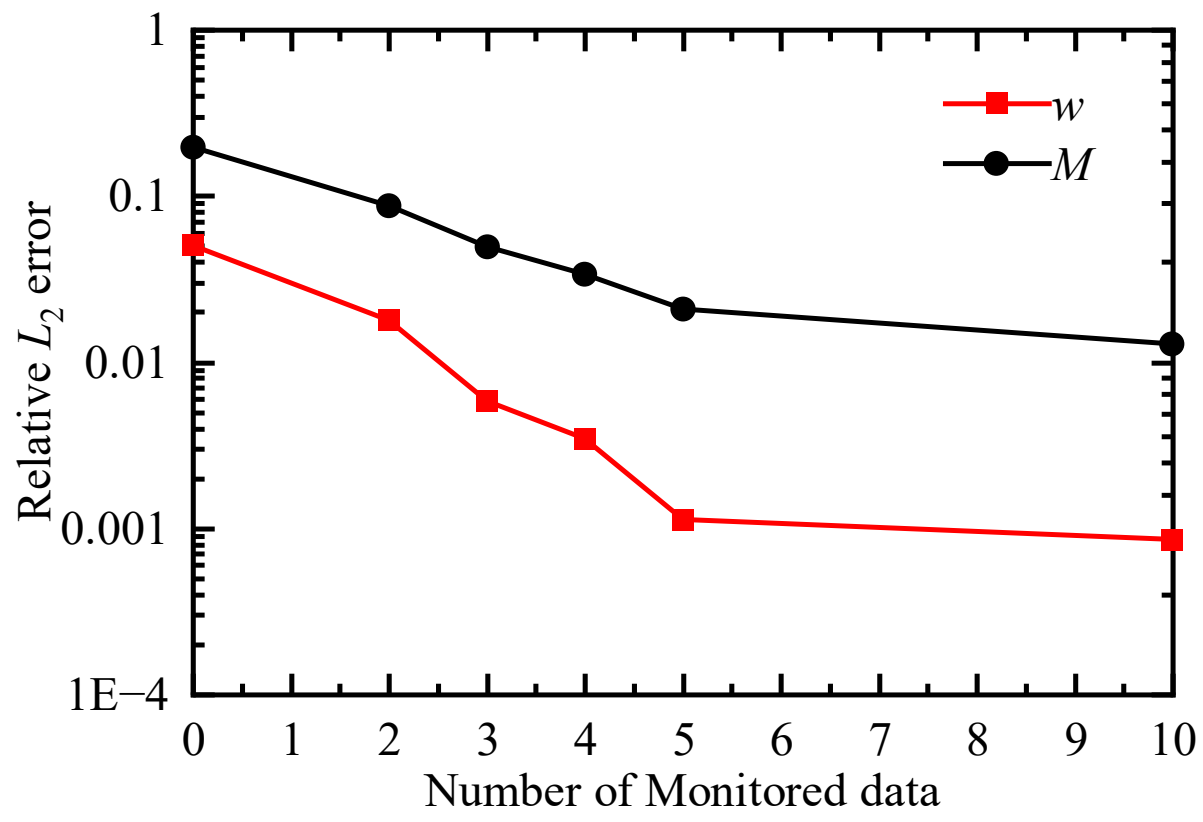

**Figure 8 Evolution of relative $L_2$ error with the data monitoring number**

## 6. Example applications of PIELM

This section shows how the PIELM method facilitates the data-driven and physics-informed analysis of tunnelling-induced soil-pile interaction using two well-documented example applications.

### *6.1. Comparison with field measurements and centrifuge model tests*

For the first example, Lee and Bassett (2007) reported a field test for tunneling-induced soil-pile interaction in London clay. In the case both ends of the pile were fixed and valuable data of pile deflection were recorded. Another example is three centrifuge model tests performed by Loganathan et al. (2000) to investigate the influence of tunneling on pile foundations in kaolin clay (pile tip and top are free). The tests were carried out at a centrifugal acceleration of 100g, with equivalent tunnel depths of 15 m, 18 m and 21 m in Test 1, Test 2 and Test 3, respectively. The corresponding input parameters for these tests are summarized in Table 8.

Based on the parametric analysis of data monitoring locations and numbers, four measured data are selected as the "monitored" data at the pile tip, pile top and near tunnelling areas. These data are incorporated into the loss function for training and are marked as solid squares, while the remaining measured data are used for validation and are marked as hollow triangles. Figure 9 and Figure 10 present the PIELM predictions against measured data in field test and centrifuge model tests. The results calculated by the purely physics-driven FDM are also plotted for comparison.

Taking the test of Lee and Bassett (2007) as an example, FDM overestimates pile deflection above the tunnel axis and underestimates it below the axis. Such discrepancies may arise from several factors, such as:

(i) simplifications in the governing equations;

(ii) idealized boundary conditions that neglect potential deflection and rotation of piles;

(iii) inaccurate quantification of tunnelling-induced external loads;

(iv) data scatter and errors in experimental tests.

By contrast, the PIELM method provides more reliable predictions of pile deformation once monitoring data are incorporated (assuming the data are accurately measured). Overall, these results underscore the value of physics-informed and data-enhanced modelling of tunnelling-induced soil-pile interactions.

**Table 8  Input parameters for physics and network**

| Parameters* | Lee and Bassett (2007) | Loganathan et al. (2000) |
|---|---|---|
| $E$ (MPa) | 30000 | 20500 |
| $v$ | 0.5 | 0.5 |
| $E_s$ (GPa) | 54 | 30 |
| $D$ (m) | 1.2 | 0.9 |
| $L$ (m) | 28 | 18 |
| $H$ (m) | 15 | 15, 18, 21 |
| $R$ (m) | 4.125 | 3 |
| $x_0$ (m) | 5.7 | 5.5 |
| $\varepsilon$ | 0.95[&] | 1 |

[&]Note: Weighted average of ground loss in the first and second excavation stages.

*$M_c$=500, $N_c$=1000, and $N_f$=2000.

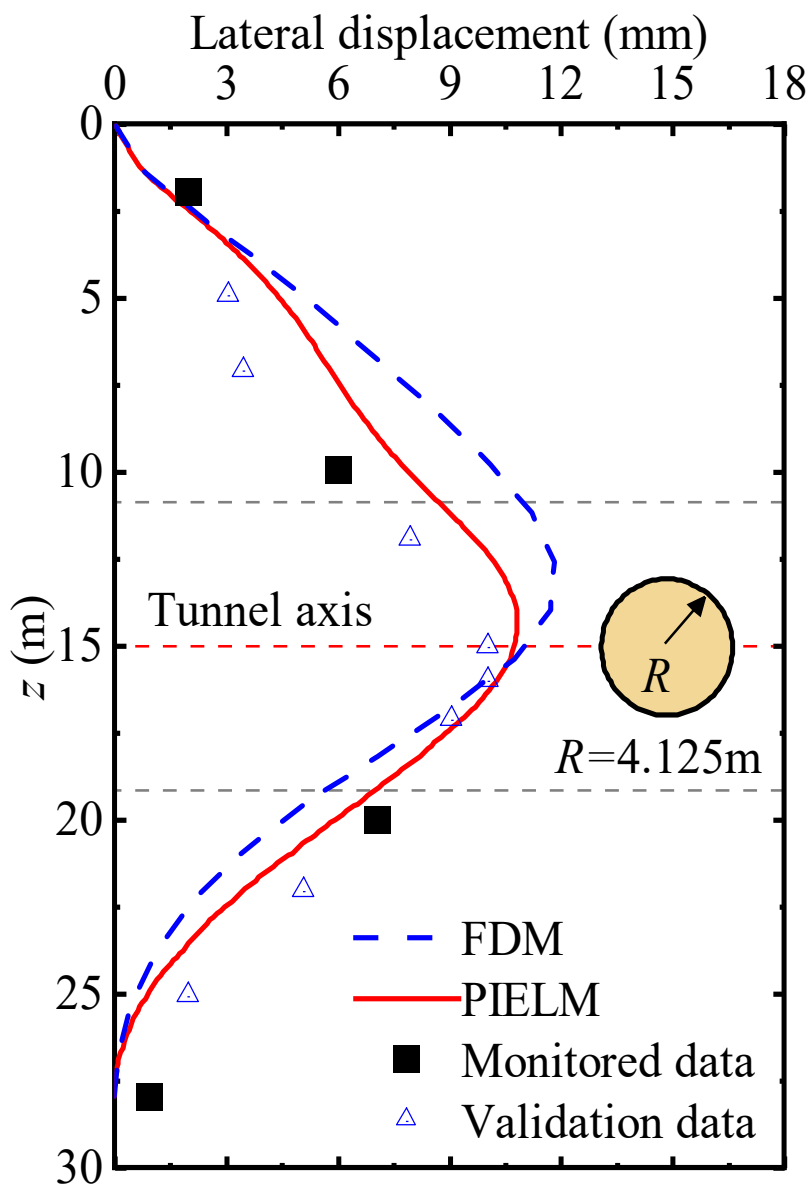

**Figure 9  Comparison with field test data of Lee and Bassett (2007)**

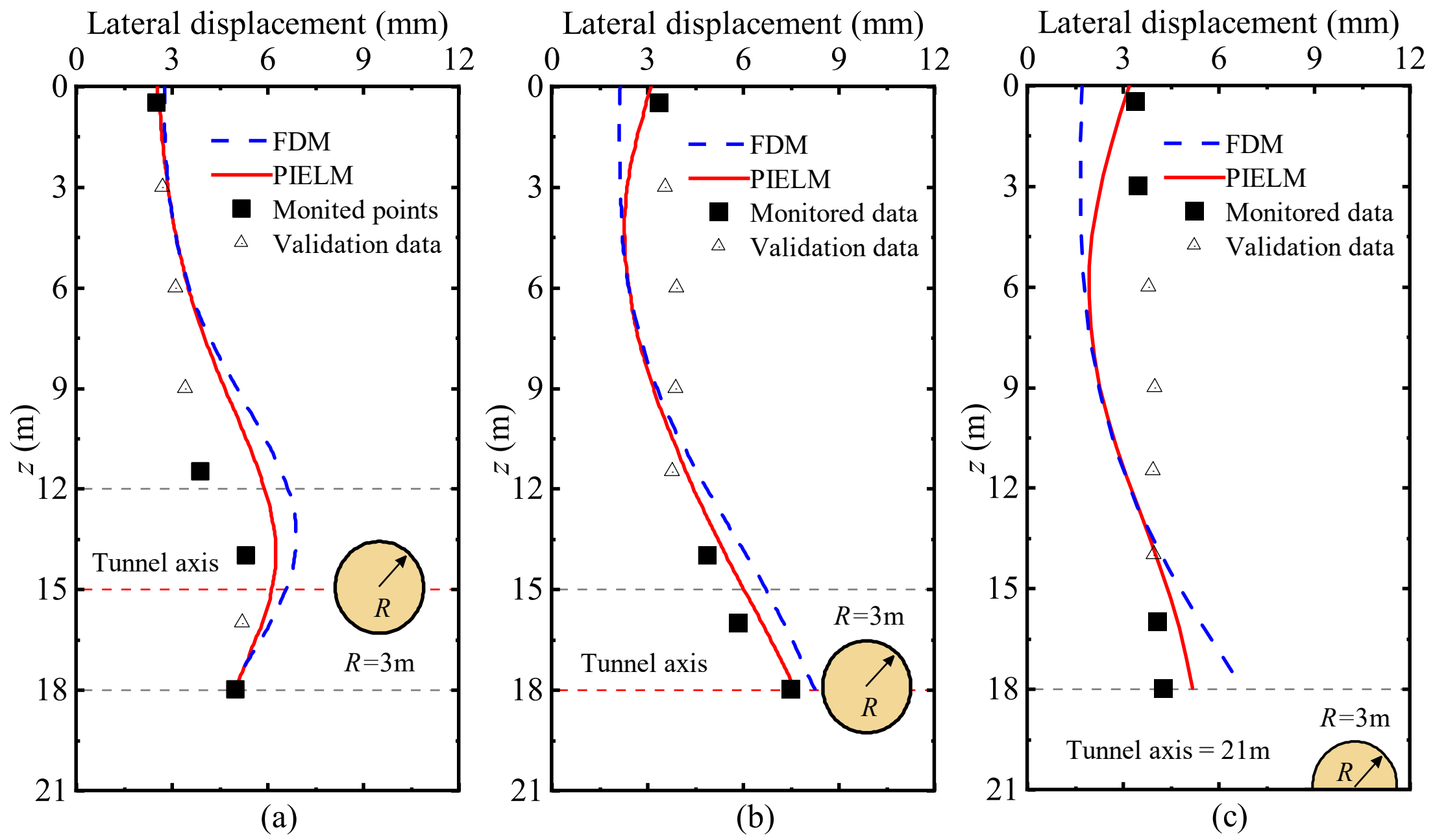

**Figure 10 Comparison with centrifuge test data of Loganathan et al. (2000): (a) Test 1; (b) Test 2; Test 3**

### *6.2. Limitations and future work*

The accuracy and efficiency of PIELM for tunneling-induced soil-pile interactions have been verified, and its potential for physics-informed and data-driven modeling has been demonstrated. Nevertheless, a few limitations remain to be addressed in future studies. First, the "monitored" data in this paper were assumed to be exact in order to rigorously evaluate the performance of PIELM, but measurement errors and even noisy data are inevitable in practice. Future extensions are necessary to quantify prediction uncertainties by integrating probabilistic modeling techniques such as Bayesian approaches (Wang and Yeung 2020; Meng et al. 2021). It is also essential to strengthen its resilience to noisy data, thereby minimizing the adverse impact of measurement. In addition, this study has confirmed that PIELM is fully applicable to real-time prediction of soil-pile interaction from a theoretical and numerical perspective, and further validation with practical monitored data is necessary in a future work.

## 7. Conclusions

This study proposes a PIELM framework for predicting tunneling-induced soil-pile interactions. The governing equations and monitored data are embedded into the ELM network through a unified loss vector. By decomposing the fourth-order pile deflection equation into a system of first-order ODEs, hard constraints are imposed in the PIELM framework to improve accuracy. PIELM trains the network by directly optimizing the loss vector using the least-squares method, which requires less than one second for network training. Verification against FDM solutions demonstrates the reliability of the approach, with relative $L_2$ errors on the order of 1e-3, 1e-2, and 1e-2 for pile deflection, bending moment, and shear force, respectively. Parametric studies yield the following key findings:

(i) When $M_c$ > 400 and $N_c$ > 100, further increases in collocation points and neurons provide limited additional improvement in accuracy.

(ii) Incorporating monitored data enhances predictive performance, but the improvement varies with their spatial distribution. The data located near the pile head/tip and tunneling zone are particularly effective.

(iii) Incorporating four to five in the critical zones (e.g., near the pile head/tip and tunneling zone) significantly improves prediction accuracy, with the given input parameters.

Finally, two application examples demonstrate the capability of PIELM for physics-informed and data-enhanced modeling of tunneling-induced soil-pile interactions. The proposed approach shows great potential for real-time prediction and warning, and in the future experimental validation and practical implementation in geotechnical engineering will be particularly helpful.

## Appendix Stiffness matrix K in FDM

The global stiffness matrix in FDM is a sparse matrix that aggregates the coefficients derived from the discrete derivatives at all nodal points. Its structure encodes both the connectivity between discrete points and the intrinsic properties of the differential equation. According to boundary conditions at the pile top and tip, the coefficients in the stiffness matrix **K** vary slightly. This appendix details **K** for three typical boundary conditions.

Rearranging Eq. (42) leads to

$$\alpha_1 w_{j+2} + \alpha_2 w_{j+1} + \alpha_3 w_j + \alpha_2 w_{j-1} + \alpha_1 w_{j-2} = f(z_j) \tag{48}$$

where the coefficients $\alpha_1$, $\alpha_2$, and $\alpha_3$ are

$$\begin{bmatrix} \alpha_1 \\ \alpha_2 \\ \alpha_3 \end{bmatrix} = \begin{bmatrix} 1 & 0 & 0 \\ -4 & -1 & 0 \\ 6 & 2 & 1 \end{bmatrix} \begin{bmatrix} \dfrac{EI}{Dl^4} & \dfrac{G}{l^2} & \dfrac{2G}{l^2} + k(z_j) \end{bmatrix} \tag{49}$$

When both the pile top and tip are free, the bending moments and shear forces at each end are zero. By substituting Eqs. (38)~(41) into Eq. (7), the pile deflection at the virtual nodes can be expressed as

$$\begin{cases} w_{-2} = w_2 - 2w_1 + 2w_0 \\ w_{-1} = -w_1 + 2w_0 \\ w_{N_f+1} = 2w_{N_f} - w_{N_f-1} \\ w_{N_f+2} = 4w_{N_f} - 4w_{N_f-1} + 2w_{N_f-2} \end{cases} \tag{50}$$

Combining the Eqs. (43), (48), (49) and (50), **K** can be obtained as

$$\mathbf{K} = \begin{bmatrix} 4\alpha_1+2\alpha_2+\alpha_3 & -4\alpha_1 & 2\alpha_1 & & & & & & \\ 2\alpha_1+\alpha_2 & -\alpha_1+\alpha_3 & \alpha_2 & \alpha_1 & & & & & \\ \alpha_1 & \alpha_2 & \alpha_3 & \alpha_2 & \alpha_1 & & & & \\ & \alpha_1 & \alpha_2 & \alpha_3 & \alpha_2 & \alpha_1 & & & \\ & & \ddots & \ddots & \ddots & \ddots & \ddots & & \\ & & & \alpha_1 & \alpha_2 & \alpha_3 & \alpha_2 & \alpha_1 & \\ & & & & \alpha_1 & \alpha_2 & \alpha_3 & \alpha_2 & \alpha_1 \\ & & & & & \alpha_1 & \alpha_2 & -\alpha_1+\alpha_3 & 2\alpha_1+\alpha_2 \\ & & & & & & 2\alpha_1 & -4\alpha_1 & 4\alpha_1+2\alpha_2+\alpha_3 \end{bmatrix}_{N_f+1} \tag{51}$$

When both the pile top and tip are fixed, the deflection and rotation angle at both ends of the pile are zero. Similarly, by substituting Eqs. (38)~(41) into Eq. (8), the expressions for the virtual nodes $w_{-2}$, $w_{-1}$, $w_{N_f+1}$ and $w_{N_f+2}$ can be obtained. Combining Eqs. (43), (48) and (49), the stiffness matrix **K** can be expressed as

$$\mathbf{K}=\begin{bmatrix} \alpha_1+\alpha_3 & \alpha_2 & \alpha_1 & & & & & & \\ \alpha_2 & \alpha_3 & \alpha_2 & \alpha_1 & & & & & \\ \alpha_1 & \alpha_2 & \alpha_3 & \alpha_2 & \alpha_1 & & & & \\ & \alpha_1 & \alpha_2 & \alpha_3 & \alpha_2 & \alpha_1 & & & \\ & & \ddots & \ddots & \ddots & \ddots & \ddots & & \\ & & & \alpha_1 & \alpha_2 & \alpha_3 & \alpha_2 & \alpha_1 & \\ & & & & \alpha_1 & \alpha_2 & \alpha_3 & \alpha_2 & \alpha_1 \\ & & & & & \alpha_1 & \alpha_2 & \alpha_3 & \alpha_2 \\ & & & & & & \alpha_1 & \alpha_2 & \alpha_1+\alpha_3 \end{bmatrix}_{N_f+1} \tag{52}$$

When the pile top is free and the pile tip is fixed, the bending moment and shear force at the top are zero, while the deflection and rotation at the tip are zero. Substituting Eqs. (38)~(41) into boundary conditions (9), we can get the expressions for the virtual nodes $w_{-2}$, $w_{-1}$, $w_{N_f+1}$ and $w_{N_f+2}$. Then the stiffness matrix **K** can be derived by combining Eqs. (43), (48) and (49):

$$\mathbf{K}=\begin{bmatrix} 4\alpha_1+2\alpha_2+\alpha_3 & -4\alpha_1 & 2\alpha_1 & & & & & & \\ 2\alpha_1+\alpha_2 & -\alpha_1+\alpha_3 & \alpha_2 & \alpha_1 & & & & & \\ \alpha_1 & \alpha_2 & \alpha_3 & \alpha_2 & \alpha_1 & & & & \\ & \alpha_1 & \alpha_2 & \alpha_3 & \alpha_2 & \alpha_1 & & & \\ & & \ddots & \ddots & \ddots & \ddots & \ddots & & \\ & & & \alpha_1 & \alpha_2 & \alpha_3 & \alpha_2 & \alpha_1 & \\ & & & & \alpha_1 & \alpha_2 & \alpha_3 & \alpha_2 & \alpha_1 \\ & & & & & \alpha_1 & \alpha_2 & \alpha_3 & \alpha_2 \\ & & & & & & \alpha_1 & \alpha_2 & \alpha_1+\alpha_3 \end{bmatrix}_{N_f+1} \tag{53}$$

## Acknowledgement

We acknowledge the funding support from Shandong Provincial Natural Science Foundation (ZR2024LZN002).

## Data Available Statement

Data are available from the corresponding author upon reasonable request.

## Conflict of Interests

The authors declare that there is no known conflict of interests